# Ambient Awareness for Agricultural Robotic Vehicles

G. Reina[1]*, A. Milella[2], R. Rouveure[3], M. Nielsen[4], R. Worst[5], M. R. Blas[6]

[1]University of Salento, Via Arnesano, 73100 Lecce, Italy
[2]CNR-ISSIA, via Amendola 122/D, 70126 Bari, Italy
[3]IRSTEA, 24 avenue des Landais, BP 50085, 63172 Aubière, France
[4]Danish Technological Institute, Forskerparken 10, 5230 Odense, Denmark
[5]Fraunhofer IAIS, Schloss Birlinghoven, 53754 Sankt Augustin, Germany
[6]CLAAS Agrosystems GmbH, CAS DK, Moellevej 11, 2990 Nivaa, Denmark

*Corresponding author. E-mail: giulio.reina@unisalento.it

**Abstract.** In the last few years, robotic technology has been increasingly employed in agriculture to develop intelligent vehicles that can improve productivity and competitiveness. Accurate and robust environmental perception is a critical requirement to address unsolved issues including safe interaction with field workers and animals, obstacle detection in controlled traffic applications, crop row guidance, surveying for variable rate applications, and situation awareness, in general, towards increased process automation. Given the variety of conditions that may be encountered in the field, no single sensor exists that can guarantee reliable results in every scenario. The development of a multi-sensory perception system to increase the ambient awareness of an agricultural vehicle operating in crop fields is the objective of the Ambient Awareness for Autonomous Agricultural Vehicles (QUAD-AV) project. Different onboard sensor technologies, namely stereovision, LIDAR, radar, and thermography, are considered. Novel methods for their combination are proposed to automatically detect obstacles and discern traversable from non-traversable areas. Experimental results, obtained in agricultural contexts, are presented showing the effectiveness of the proposed methods.

**Nomenclature**

| | |
|---|---|
| QUAD-AV | Ambient awareness for autonomous agricultural vehicles |
| HDR | High dynamic range |
| FMCW | Frequency modulated continuous wave |
| CFAR | Constant false alarm rate |
| EM | Expectation maximization |
| GMM | Gaussian mixture models |
| $x$ | Sensor observation |
| M($x$) | A posteriori class likelihood |
| P | Positive predictive value |
| RP | Negative predictive value |
| CUDA | Compute unified device architecture |

# 1. Introduction

Agricultural mechanisation was recognised as the 7th greatest engineering achievement of the 20th century by the US National Academy of Engineering (Constable and Somerville, 2003). This testifies the importance of farm machinery that turned traditional agriculture to an industry through increased efficiency and productivity. Next expected step is from mechanisation to automation, as one of the possible engineering goals of the 21st century. In this respect, the last few years have seen an increasing technological transfer from robotics to agriculture towards the development of intelligent vehicles (Tobe, 2015) that can lead to the saving of labour time (see e.g., Auat Cheein and Carelli, 2013), and more efficient farming methods, including regular monitoring of plant growth and precision type of plant nursing (see e.g., Klose et al., 2010). One specific objective is that of increasing the level of driving automation of tractors and implements in crop fields, as recognised by the Horizon 2020 Multi-Annual Roadmap for Robotics in Europe (HORIZON 2020, 2015). This is especially important considering the latest trend in agriculture of using a group of cooperating vehicles to further improve efficiency and productivity (Reinecke et al., 2013).

Although auto-guided agricultural vehicles using GPS-based navigation systems have been in practical use for some years, these systems do not provide any information about the "dynamics" of the environment. For instance, in most agricultural applications, a coarse map of the terrain where the vehicle operates is often available. However, we cannot blindly trust this map, because it might contain errors due to recent changes in the field caused by humans or by nature. Furthermore, the map does not contain any information about moving objects (e.g., human beings, animals, and vehicles) that may be present in the scene and it requires correction data as well as an extremely accurate knowledge of crop row locations. This causes a major safety issue related to any type of autonomous navigation and operation. Therefore, a large body of research has been recently devoted to guiding agricultural robots within crop rows using local sensors that can provide a rich source of information including colour, texture and 3D structure (English et al., 2014).

The challenge is now to develop smarter, semi- or fully-autonomous vehicles that can operate safely in semi-structured or unstructured dynamic environments, in which humans, animals and other machines may be present. To address this issue, researchers need to endow the vehicle with a clear understanding of its surrounding world. The cognitive ability to perceive the environment is, in many cases, a matter of guaranteeing the safety of people, animals and risking expensive machines or causing yield damage.

The development of a perception system to increase environment awareness of an agricultural vehicle is the focus of the Ambient Awareness for Autonomous Agricultural Vehicles (QUAD-AV) project. The overall goal of the QUAD-AV project is the implementation of a multi-sensor platform and sensor processing algorithms for driving assistance/automation that can be integrated on-board an agricultural vehicle, such as a tractor or harvester, to improve and facilitate its safety. Ambient awareness within crops is essential to implement many tasks, including obstacle detection and traversability assessment.

Reliable obstacle detection in agricultural settings is challenging due to the complex unstructured nature of such environments. Fig. 1 shows four typical types of hazards that can be encountered in agricultural settings (Farm Equipment Accidents, 2014): positive obstacles (e.g., trees, metallic poles, buildings), negative obstacles (e.g., holes, ditches, etc.), moving obstacles (e.g., vehicles, people and animals), and difficult terrain (e.g., significant slope, water, etc.). Obstacles may also vary greatly from situation to situation, depending on type of crop, fruit, vegetable or plant grown, curvature of landscape, etc. Current agricultural vehicles have no, or at most very poor, understanding of such obstacles.

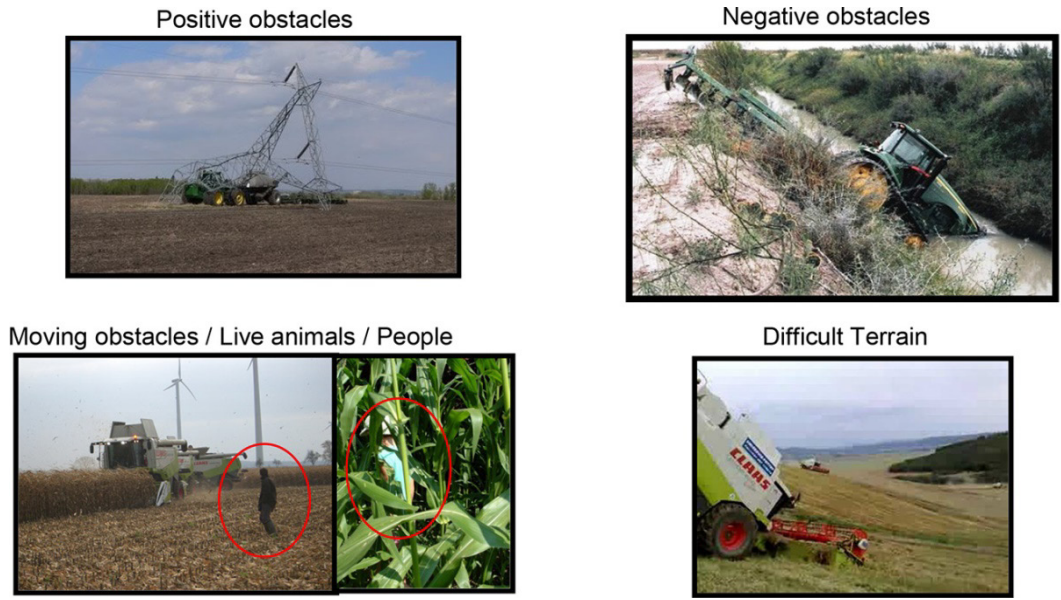

**Fig. 1** - Typical obstacles that can be encountered in agricultural settings (Farm Equipment Accidents, 2014).

Closely related is the ability to discern traversable from non-traversable regions. Solutions based on ground plane estimation typically adopted in indoor or structured outdoor environments (e.g., on roads) are not applicable in agricultural contexts, as the soil surface may be rutted and covered by crop plants with variable size and shape. Weather conditions and environmental factors like dust and smoke may further modify the appearance of ground and obstacles therein. Due to such a variety of situations and problems that may be encountered, no sensor modality exists that can guarantee reliable results in every scenario. Any candidate sensor has its strengths and drawbacks. Therefore, a complementary sensor suite should be used to gain the best overall performance.

In the context of this research, four technologies are investigated: stereovision, LIDAR, radar, and thermography. State-of-the-art sensors, some of which custom-built by the partners are modified and combined in order to be demonstrated in an agricultural context. Novel sensor processing methods for obstacle detection and traversable ground identification are developed and experimentally validated through field tests. An underlying theme of the proposed methods is the reliance on statistical frameworks for automatic learning and segmentation of obstacle/ground classes.

This paper presents an overview of the perception methods developed within the QUAD-AV project, including two novel solutions for combining stereovision with radar and thermography, respectively. In detail, after a review of the state of the art in section 2 and a description of the system architecture in section 3, a ground classifier using 3D point clouds reconstructed by a multi-baseline stereovision system is discussed in section 4. It is shown that it can work in a wider range of distances than standard stereovision. In section 5, a statistical framework to combine stereovision and LIDAR for traversable ground detection is introduced, showing that the combined system improves the outcome of each single sensor. Successively, an original obstacle detection approach combining radar and stereovision is proposed in section 6. The radar ensures accurate and fast 2D obstacle detection, whereas stereovision is employed to augment the output of the radar-based obstacle detector with 3D and colour information. Finally, section 7 describes a novel integration of high dynamic range (HDR) stereovision and thermal imaging, which aims to improve the detection of particular obstacles, such as humans and animals. The conclusion summarises the lessons learned and the follow-up actions.

Note that, although the farm operations supported by QUAD-AV methodology include any precision application that relies on the accuracy of the local sensors, including obstacle avoidance and auto-guidance, the QUAD-AV does not supply any application technology, but does supply the data to inform such technologies.

## 2. Related Work

Future agricultural vehicles will be required to operate throughout extensive fields, such as broad-acre crops, with limited human supervision, while preserving safety. This will primarily need the development of robust navigation systems (Ojeda et al., 2006), including obstacle detection and traversability assessment methods (Kise and Zhang, 2008).

Research in these contexts includes methods that aim to explicitly model and detect different types of obstacles, and then determine the free space as the traversable ground. For instance, in (Rankin et al., 2005), positive/negative obstacles, excessive slopes, water obstacles, non-traversable tree trunks and low-overhang obstacles, are explicitly modelled and identified in the scene using stereovision. An intrinsic drawback of such approaches is that only a limited number of obstacle types may be modelled and identified. This problem is overcome in novelty or change detection methods, where obstacles are defined as regions of the image that are dissimilar to recent views of the world. An example of such approaches in agriculture can be found in Ross et al. (2014), where a novelty detector estimates the probability density in image descriptor space and incorporates image-space positional understanding to identify potential regions for obstacle detection via dense stereo matching.

Alternatively, approaches that build a model of traversable ground surface and detect the regions in the surroundings of the vehicle that respond to that model have been proposed. These methods are advantageous in that they can be used to identify different terrain types and set the vehicle's control system accordingly. In general, depending on the adopted learning strategy to model the terrain surface, ground identification algorithms can be subdivided into deterministic (i.e., no learning), supervised and self-supervised methods. Deterministic techniques assume that the characteristics of the ground are fixed, and therefore they cannot easily adapt to changing environments (Huertas et al., 2005; Singh et al., 2000). Without learning, such systems are constrained to a limited range of predefined environments. A number of systems that incorporate supervised learning methods have been recently proposed, many of them in the automotive field and for structured environments (e.g., in road-following applications) (Manduchi et al., 2003; Rasmussen, 2002). For instance, in Rasmussen (2002), features from a laser range-finder and colour and texture image cues are used to segment ill-structured dirt, gravel, and asphalt roads by training separate neural networks on labelled feature vectors clustered by road type. These systems were trained offline using hand-labelled data, thus limiting the scope of their expertise to environments seen during training.

Only recently, self-supervised systems have been developed that reduce or eliminate the need for hand-labelled training data, thus gaining flexibility in unknown environments. With self-supervision, a reliable module that determines traversability can provide labels for inputs to another classifier. A self-supervised framework that predicts the mechanical properties of distant terrain based on a previously learned association with visual appearance is proposed in (Brooks and Iagnemma, 2012). Self-supervised learning also helped win the 2005 defense advanced research projects agency (DARPA) Grand Challenge: the winning approach used a probabilistic model to identify road surface based on colour information extracted immediately ahead of the vehicle as it drives (Dahlkamp et al., 2006).

In this paper, a self-learning classifier, previously presented by the authors in (Milella and Reina, 2014), is briefly recalled, which aims to label data from a 3D point cloud as ground or non-ground based on geometric features. This approach does not entail ground plane reasoning, nor does it require a priori knowledge of the environment and of the obstacles therein, thus allowing the system to automatically adapt to terrain and environmental changes. The classifier can be applied to data acquired by any 3D range sensor. Here, it is demonstrated for scene segmentation from the near range up to several meters away from the vehicle, using 3D points reconstructed by a multi-baseline stereovision system.

Although stereovision provides a rich source of information for obstacle detection and characterization, pure vision-based solutions are prone to fail in poor lighting conditions or in the presence of visual obscurants like dust, smoke or fog, which are all likely to occur in agricultural settings. Stereovision data should be therefore combined with other sensors featuring complementary characteristics.

For example, LIDAR-stereovision integration has been proposed by many. Two main methodologies can be found: a priori and a posteriori integration. The first approach combines ranges obtained from the two sensors at raw data or feature level (Badino et al., 2011; Aycard et al., 2011). In contrast, *a posteriori* methods aim to combine the (higher level) output of both sensors to enhance the final response of the system (Dima et al., 2004). As the sensor fusion is performed at a decisional level rather than at data or feature level, it is independent of the types of sensors used. The approach proposed in this work can be considered as an *a posteriori* method. The main novelty relies on the adoption of an adaptive self-learning scheme to combine statistically the results of individual LIDAR- and stereovision-based classifiers to produce a unique classification result. This mitigates the drawbacks that would result in using each sensor modality separately.

However, vision and LIDAR are both affected by weather phenomena or other environmental factors, such as smoke and dust. In contrast, radar penetrates dust and other visual obscurants (Peynot et al., 2009). Nevertheless, radar has shortcomings as well, like specularity effects and limited range resolution, which may result in poor environment survey or make it difficult to extract object features for classification and scene interpretation tasks. Due to the complementary characteristics of the radar and vision sensors, it is reasonable to combine them in order to get improved performance. Radar-stereovision combination for off-road applications represents a novel contribution as most radar and vision fusion algorithms in the literature refer to on-road automotive systems (Ji and Prokhorov, 2008; Wu et al., 2009), under the two main assumptions of planar ground and car-like obstacles. Few examples are provided for off-road applications. For instance, in (Milella et al., 2011; Milella et al., 2015) radar sensing is used to supervise a monocular system in a rural environment.

As an alternative or complementary sensor, recent investigations have demonstrated the use of thermal cameras in challenging environmental conditions. For instance in Brunner et al. (2013), it is shown that thermal imaging is useful to complement vision for vehicle navigation in adverse conditions including smoke, fire, and night time. Since it does not require any light source, while it relies only on the thermal radiation emitted from every object, a thermal camera is an attractive solution to detect humans and animals, also when they are hidden or obscured in cluttered environments. In the context of QUAD-AV, thermal imaging is combined with HDR stereovision for obstacle detection in agricultural fields to overcome problems with shade and direct sunlight.

3. Multi-sensory Perception

The potential of four technologies is investigated: stereovision, LIDAR, radar, and thermography. Advantages and disadvantages of each of these technologies are summarized in Table 1, whereas Table 2 collects the technical details and the approximate cost of the sensor suite. Note that while the costs refer to standalone equipment, significant savings may be achieved by volume purchase of board level sensors (e.g., a cellphone thermal camera FLIR ONE costs about 150 €) that may be weatherproofed using a common sensor housing.

The multi-sensor system was integrated with two experimental test beds for field demonstration. The first data acquisition campaign was conducted in a farm facility near Helsingør, Denmark. Fig. 2(a) shows the positioning of the sensors on the CLAAS AXION 840 tractor. The second experimental campaign was performed using an off-road land vehicle (Fig. 2(b)) that was made available by the partner National Research Institute of Science and Technology for Environment and Agriculture (IRSTEA) at the farm facility of Montoldre, France. In both experiments, various typical agricultural scenarios were encountered including positive obstacles (high-vegetated areas, trees, crops, metallic poles, buildings, agricultural equipment), negative obstacles (ditches and other depressions), moving obstacles (vehicles, people and animals), and difficult terrain (steep slopes, highly irregular terrain, etc.). The vehicle was manually driven, as the onboard sensors acquired data from the surrounding environment. Then, the proposed algorithms for obstacle detection and traversability assessment were applied offline.

**Table 1 -** Advantages and disadvantages of different sensor modalities for outdoor perception

| Sensor modality | Advantages | Disadvantages |
|---|---|---|
| Mono/Stereovision | Natural interpretation for humans<br>Relatively high resolution<br>Relatively high sampling rate<br>Rich content (colour and texture for mono and stereovision, range for stereovision) | Risk of occlusions<br>Sensitive to lighting conditions<br>Poor performance in low visibility conditions (rain, fog, smoke, etc.)<br>Relatively short range (up to 15-30 m)<br>Advanced processing algorithms and special hardware for data acquisition. |
| LIDAR | Accurate ranging over medium range (up to 30-40 m)<br>Narrow beam spread<br>Fast operation<br>Fast lock on time<br>Integration with rotating platforms for 3D acquisition | Costs related to accuracy and range<br>Sensitive to dust, fog and rain, and rounded surfaces<br>Some risk of occlusion<br>No colour or texture information<br>Relatively low sampling rate (except for high-end sensors) |
| Radar | Robustness to environmental conditions<br>Long range (up to 100-150 m)<br>Panoramic perception (360°)<br>Multiple targets | Difficulty in signal processing and interpretation<br>Low resolution<br>3D map limited to pencil beam radar |
| Thermography | Invariant to illumination<br>Robust against dust and rain<br>Detect humans and animals | Relatively low resolution<br>Some risk of occlusion<br>Difficulty in calibration |

**Table 2 -** Technical details of the sensor suite including the approximate costs

| Sensor | Type/model | FOV | Range | FPS | Cost |
|---|---|---|---|---|---|
| Stereovision | Multi-baseline (12/24 cm, and 40/80 cm) and multi-focal system (3.8/12 mm), Bumblebee XB3 plus Flea3 cameras (Milella and Reina, 2014) | Bumblebee XB3: 66 deg(H) x 50deg(V)<br><br>Flea3 cameras: 30deg(H) x 23deg(V) | Bumblebee XB3: 2-30m<br><br>Flea3 cameras: 6-60 m | 1-10 Hz | ~3 k€<br><br>~0.7 k€ |
| LIDAR | Rotating rangefinder, 3DSL with SICK LMS111 (Reina et. 2013) | 360deg(H)x 270deg(V) | 0.5 to 17 m | 0.3 Hz | ~6k€ |
| Radar | Mechanically scanned, 24 GHz, custom-built, K2Pi | 360deg(H)x25deg(V) | 3-100 m | 1 Hz | n/a* |
| Thermography | Flir a615 thermal camera | 40deg(H)x30deg(V) | 2-20 m | 20 Hz | ~11 k€** |
| * Although the radar is custom built at IRSTEA, it is similar in performance to other commercially available systems. For example, http://www.nav-tech.com/Industrial%20Sensors2.htm<br>**Note that the application developed in this research may be implemented using lower grade sensors, e.g. Flir A25 (~3 k€) | | | | | |

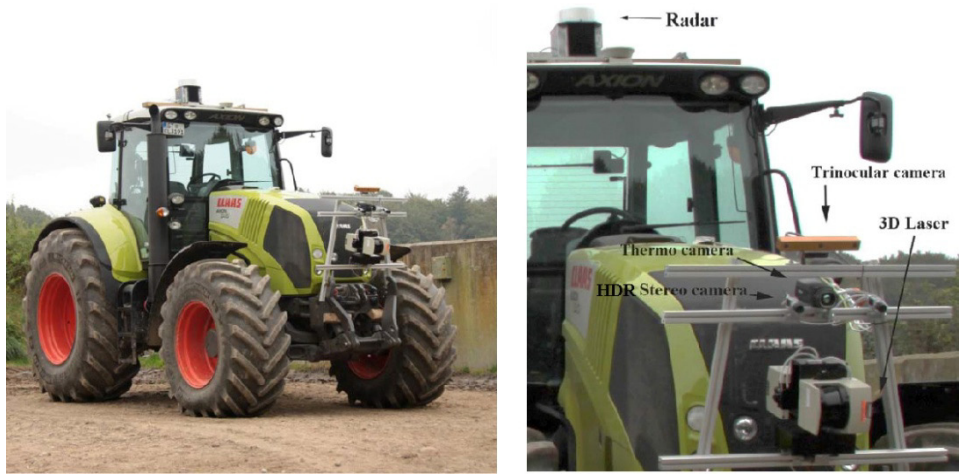

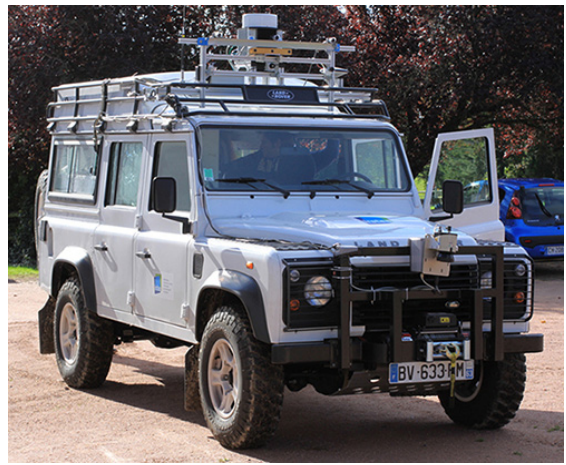

**Fig. 2 -** Test beds used for field validation along with their sensor suites. (a) CLAAS AXION 840 tractor. (b) Off-road land vehicle.

## 4. Multi-baseline Stereovision

Stereovision is a widely adopted input for outdoor navigation, as it provides an effective technique to extract range information and perform complex scene understanding tasks (Hanawa et al., 2012; Reina and Milella, 2012). Nevertheless, the accuracy of stereo reconstruction is generally affected by various design parameters, such as the baseline. On one hand, a larger baseline guarantees higher accuracy at each visible distance. On the other hand, a long baseline leads to a loss of information in the near range; furthermore, it requires a larger disparity search range, which implies a greater possibility of false matches. Hence, the choice of the optimal baseline results from the balance of opposing factors, depending on the requirements of the target application.

In this research, a multi-baseline stereovision rig was developed, which allows an off-road vehicle to perform accurate 3D scene reconstruction and segmentation in a wide range of distances. The stereo frame is shown in Fig. 3(a).

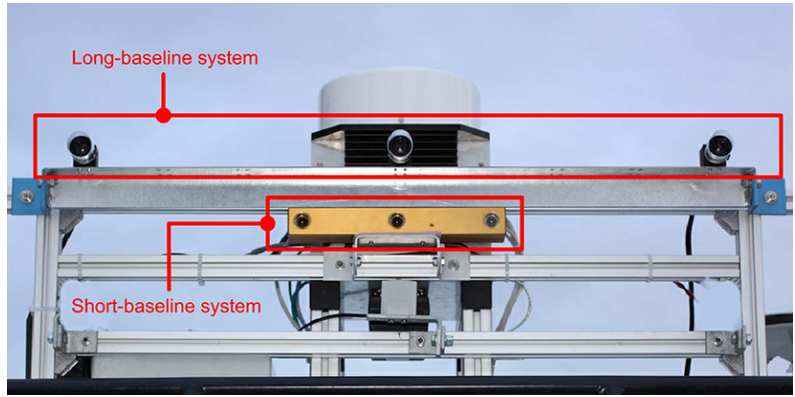

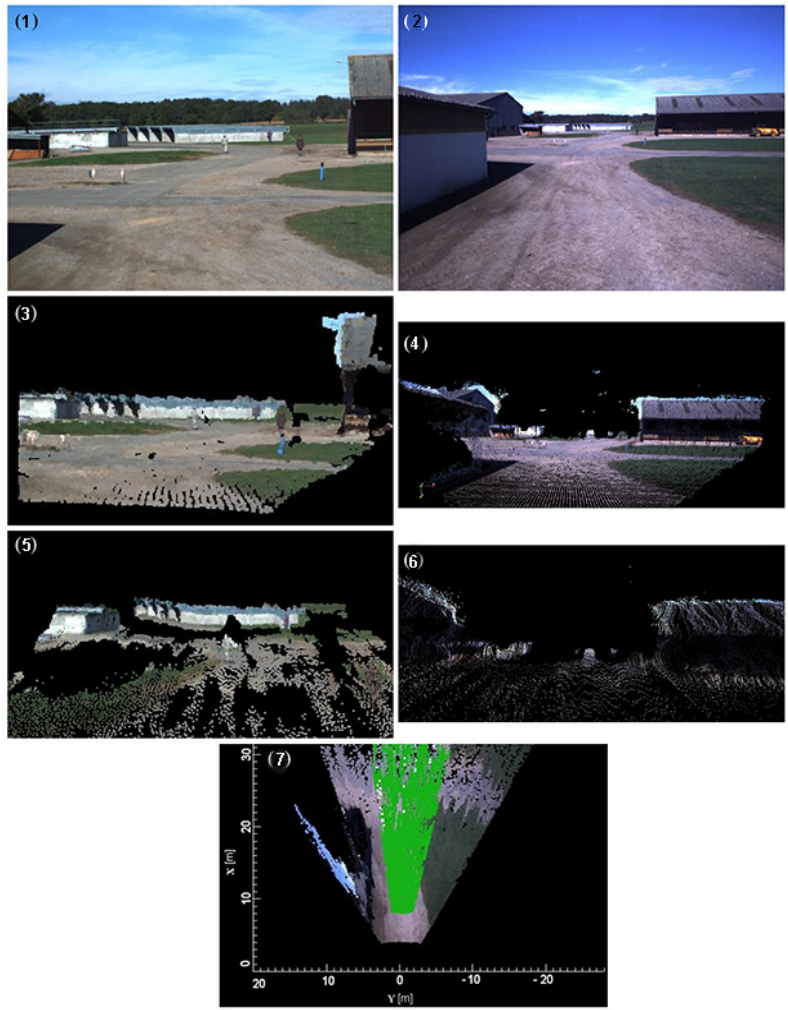

**Fig. 3 -** (a) Multi-baseline stereovision rig. (b) Sample scenario acquired during field experiments in a farm: 1) original, reference image as obtained from the long-baseline (Flea3) system, 2) original, reference image as obtained from the short-baseline system (XB3), 3) point cloud reconstructed by the long-baseline module for the whole scene, 4) point cloud reconstructed by the short-baseline module for the whole scene, 5) close-up of far regions in the long-baseline reconstruction, 6) close-up of far regions in the short-baseline reconstruction, 7) upper view of the near regions: RGB points are used for the short-baseline and green points for the long-baseline. It is shown that the long-baseline system provides accurate information in the distance while losing information in the vicinity of the vehicle compared with the short-baseline camera.

It is composed by two trinocular heads, one featuring a short-baseline system and the other one featuring a long-baseline system. The short-baseline camera is the Bumblebee XB3 (Point Grey, Richmond, BC). It consists of a trinocular stereo head with 3.8mm focal length lenses, with two stereo configurations: a narrow stereo pair with a baseline of 0.12m using the right and middle cameras, and a wide stereo pair with a baseline of 0.24 m using the left and right cameras. The long-baseline trinocular system is custom built. It comprises three Flea3 cameras (Point Grey, Richmond, BC) with 12.0 mm focal length lenses, disposed in line on an aluminium bar to form two baselines: a narrow baseline of 0.40 m using the left and middle cameras and a wide baseline of 0.80 m, using the left and right cameras.

The two trinocular cameras can be either used simultaneously to widen the overall perception range of the vehicle, or alternately depending on the vehicle travel conditions. For instance, the short-baseline configuration is useful in low-speed operations, where less noisy measurements are needed, while the long-baseline is suitable when the vehicle travels at higher speed, enabling it to perceive far away obstacles. In addition, the long-baseline can improve the quality of the stereovision data for distant terrain mapping. The 3D reconstruction in the short and long range is shown in Fig. 3(b) for a scenario with relatively flat ground and a building on the left in the vicinity of the vehicle, and buildings and people in the far range. It can be observed that the farthest building is filtered out in the reconstruction provided by the short-baseline camera, while it is reconstructed by the long-baseline system. Conversely, the long-baseline camera is not able to detect nearby regions, as its point of view is located farther than the one of the short-baseline camera. In addition, it has a narrower angular field of view that causes the loss of important information on the building on the left of the vehicle, which is detected by the short-baseline system, instead.

### 4.1. Geometry-based obstacle detection

The 3D point cloud returned by either trinocular camera provides a rich source of information for the vehicle to perform key navigation tasks, such as terrain estimation. A geometry-based classifier is proposed to label data from a 3D point cloud as ground (i.e., traversable area) or non-ground (i.e., non-traversable area) according to their geometric properties. It adopts a self-learning scheme, whereby training instances to build the ground model are automatically produced using a rolling training set. The latter is initialised at the beginning of robot operation via a bootstrapping approach. No a priori assumption about the terrain surface characteristics is needed. The only hypothesis to initialise the training set is that the system starts its operation from an area free of obstacles in proximity of the vehicle, so that the sensor initially "looks" at ground only. Then, the point cloud is processed to get a set of features, representative of their respective geometric properties. To decrease the computational burden, the number of points is reduced using voxelisation. A 3D voxel grid is created over the input point cloud space, and all the points in each voxel are approximated with their centroid. Then, the point cloud is divided into a grid of terrain patches projected onto a horizontal plane. Finally, geometric features are extracted as statistics obtained from the point coordinates associated with each terrain patch.

When sufficient data is accumulated, the geometry-based ground classifier is trained, and the ground class is related with the point cloud properties. This allows the system to predict the presence of ground in successive scenes, using a Mahalanobis distance-based classifier (Duda et al., 2001). To account for variations in ground characteristics during the vehicle travel, the ground model (i.e., the training set) is continuously updated using the most recent acquisitions. The reader is referred to Milella and Reina (2014) for a detailed description of the single steps of the approach.

Some typical results of the classifier obtained in a maize crop are shown in Fig. 4. Pixels associated with ground-labelled cells are denoted in green, whereas 3D points falling into cells labelled as non-ground are marked using red dots.

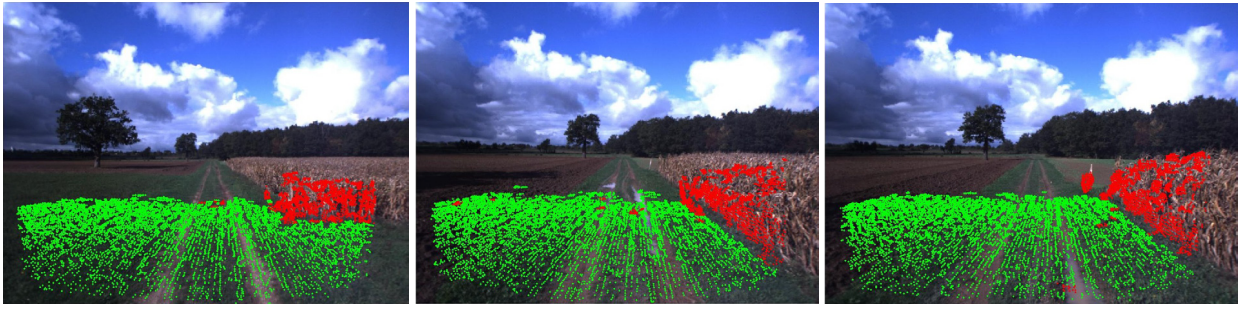

**Fig. 4 -** Results obtained from the geometry-based classifier in a maize crop. Green dots denote ground-labelled points, red dots denote points classified as non-ground.

To provide a quantitative evaluation of the classification algorithm, precision, recall (i.e., true positive rate), specificity (i.e., true negative rate), accuracy, and F1-score were measured for a subset of salient images (i.e., $s_b$=135 for each camera) taken from different data sets acquired by both the short-baseline and the long-baseline trinocular sensors. This subset was hand labelled to identify the ground truth corresponding to each pixel. Sample images from this test set are shown in Fig. 5, for the long-baseline (left) and the short-baseline (right), with overlaid the classification results. The numerical results obtained on the whole subset are reported in Table 3. It can be seen that both systems achieve good classification performance with accuracy of 86.5 and 88.8%. It is worth to note that the proposed classifier is of general validity and can be used with any 3D range sensor, as will be shown for LIDAR-stereovision combination.

**Table 3 -** Performance of the individual long-baseline (Flea3) and short-baseline (XB3) classifiers, through comparison with ground-truth data obtained by manual labelling.

| Camera | Precision (%) | Recall (%) | Specificity (%) | Accuracy (%) | F1-score (%) |
|---|---|---|---|---|---|
| Flea3 | 94.0 | 84.6 | 90.1 | 86.5 | 89.0 |
| XB3 | 95.2 | 90.6 | 81.1 | 88.8 | 92.9 |

## 5. LIDAR-Stereovision combination

LIDAR and vision systems are complementary in many ways: for example, when performing a three-dimensional reconstruction of the environment, LIDAR offers accurate, yet sparse, data, whereas vision provides dense, but less accurate measurements. Furthermore, scanning LIDARs may feature low sampling rates (0.1-1 Hz) resulting in difficulties to detect dynamic obstacles. Expensive LIDARs like Velodyne (www.velodyne.com/LIDAR) are an exception. In addition, LIDAR may also result in limited sensing range due to the particular sensor location onboard the vehicle. This is the case for the configuration used in this research (please refer to Fig. 2), where the LIDAR features less range than the stereovision system because of the low height from the ground that limits the look-ahead distance. In contrast, images from a stereo camera can be processed at a fast frame rate allowing static and dynamic obstacles to be detected. This can be particularly useful for reactive planning responding to dynamic obstacles.

This paper proposes a statistical framework to combine LIDAR and stereovision for obstacle detection purposes, as preliminary introduced in Reina et al. (2013). Specifically, the geometry-based classifier proposed in Milella and Reina (2014) is, first, applied to LIDAR and stereovision independently. Afterwards, the outputs obtained from the two single-sensor classifiers are statistically combined to improve the overall perception ability of the system. The combination of the two sensor modalities proceeds as follows. Assume that we are given a set of classifiers, which have already been trained to provide as output the class *a posteriori* likelihood in the form of the Mahalanobis distance from the class centre. In detail, given an unknown observation x, the classifier *i* produces estimates of the *a posteriori* class likelihood, that is $M_i(x)$. In our case *i* = L, S, where L stands for LIDAR-based classifier, and S for stereo-based classifier. Our goal is to devise a way to come up with an improved estimate of a final *a posteriori* likelihood M(x) based on all the resulting estimates from the individual classifiers.

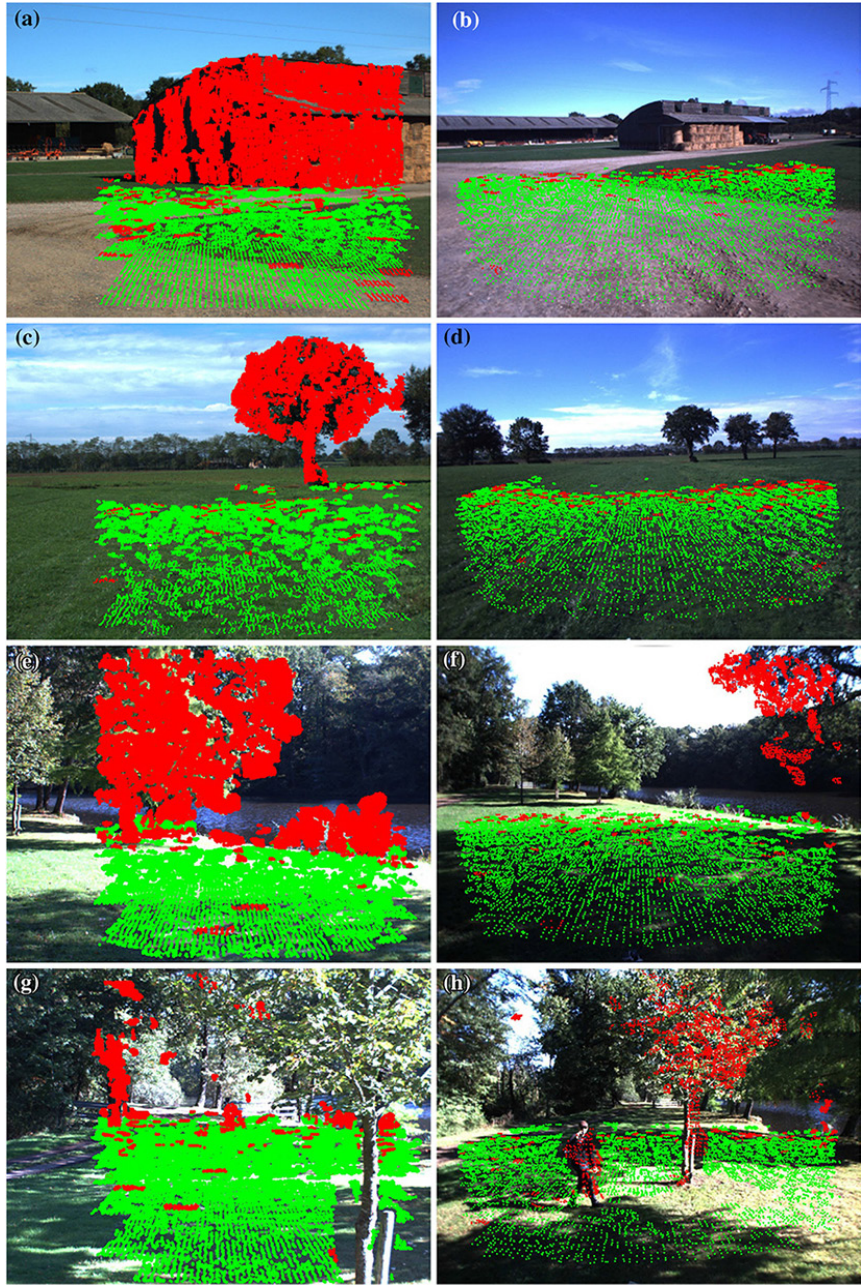

**Fig. 5 -** Results obtained from the geometry-based classifier using multi-baseline stereovision for some salient images with different terrain types: i.e., unpaved road (a, b), low grass (c–h) and obstacles (i.e., buildings (a), trees and bushes (c, e, g, h), people (h)). Left: classification of far regions by the long-baseline system. Right: classification of near regions by the short-baseline system. Green dots denote ground-labelled points, red dots denote points classified as non-ground.

One way is to weight the individual output obtained from the classifiers with their prior probabilities that can be statistically quantified using ground-truth data. This analysis would provide various statistical quantities including for instance a confusion matrix. By appropriately normalising true positives (TP), false positives (FP), true negatives (TN), false negatives (FN), obtained from the confusion matrix, we can construct (empirical) expected rates of positive predictive value or precision (P) and negative predictive value or rejection precision (RP),

$$P = \frac{TP}{TP+FP} \quad (1)$$

$$RP = \frac{TN}{TN+FN} \quad (2)$$

Being normalised, these rates are also probabilities. So now, we have values of uncertainty associated with the LIDAR and stereovision only ground detection in the form of prior probabilities over these detections. These (prior) probabilities can be used as weights to combine statistically the decision of each classifier through a weighted sum and obtain a unique classification result,

$$M(x) = \sum_{i=L}^{S} \frac{W_i M_i(x)}{W_L + W_S} \qquad (3)$$

where the weight, $W_i$, is equal to $P_i$ or $RP_i$ if the observation *x* is labelled as ground or non-ground, respectively, by the classifier *i*. As a result, the uncertainty associated with each classifier is propagated throughout the final classification result.

It is worth to note that the proposed data fusion framework is of the "weak" type, i.e., data are combined after each sensor has produced high-level information. Indeed, stereovision and LIDAR produce compatible data (i.e., range measurements), which would make it feasible a "strong" data fusion approach, where data are fused at a low level, i.e., before high-level information has been produced. However, in this work, a high level fusion scheme was preferred, to keep the system independent from the data types and potentially applicable to any sensor combination although they differ in accuracy and sampling rate, which is the case for the multi-sensory perception system that serve advanced driver assistance systems. The result of the stereo-LIDAR combination is shown in Fig. 6 for a sample case. The first row shows, from left to right, the original visual image overlaid with the points reconstructed by LIDAR, stereo and LIDAR-stereo fusion, respectively. Green dots denote points belonging to ground-labelled cells, whereas red dots are used to indicate pixels falling into non ground-labelled cells. The second row shows, from left to right, the traversability map as obtained using only LIDAR data, only stereovision data and the combination of both.

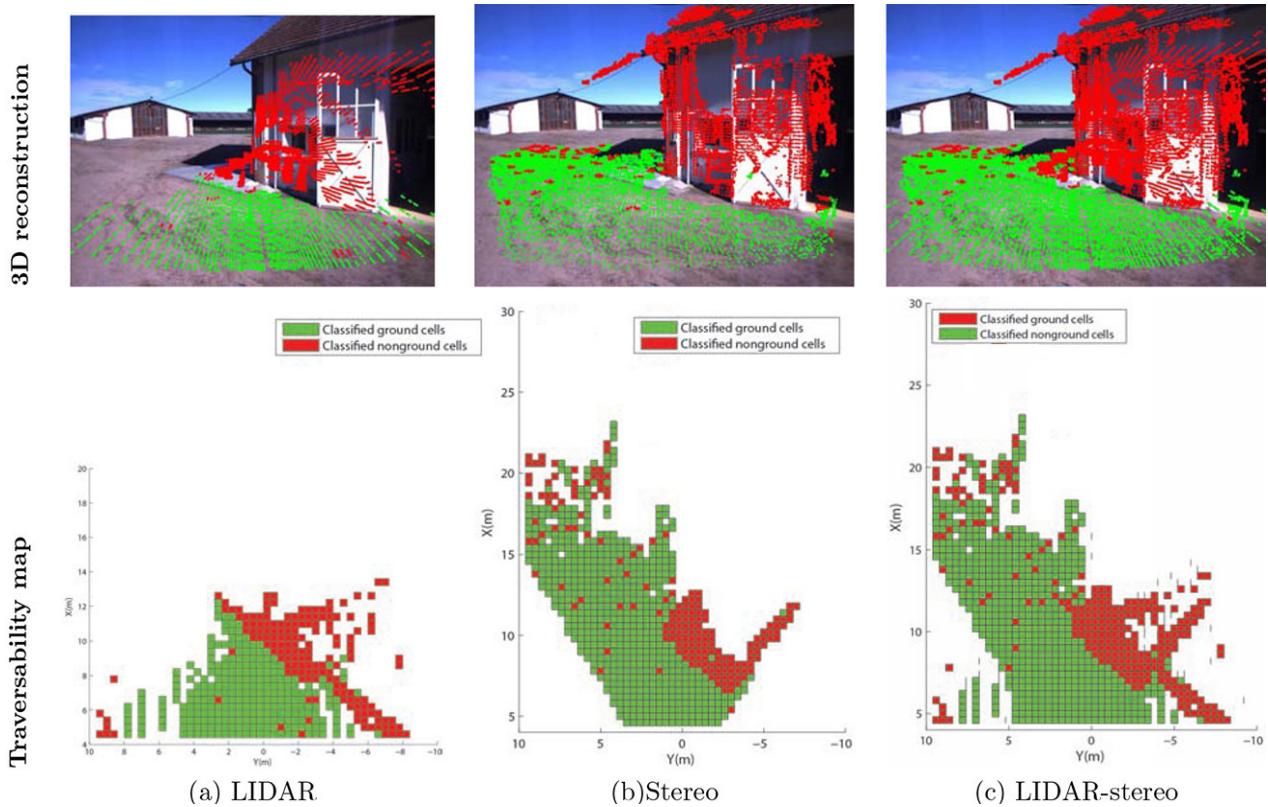

**Fig. 6 -** Multi-sensor ground estimation: (a) LIDAR classifier, (b) Stereovision classifier, (c) LIDAR-stereovision classifier.

Green cells denote obstacle-free patches, whereas red cells denote cells containing obstacles, and thus non-traversable.

A quantitative evaluation of system performance was obtained over a subset of salient images (i.e.,

$s_b$=113) taken from various data sets and referring to significant scenarios. This subset was manually labelled to gain reference ground-truth data. The combined LIDAR-stereovision classifier was compared against the LIDAR and stereovision classifier using only those cells that were observed by both sensors. For the selected frames, the true ground patches amounted to a total of 24,953, whereas the true non-ground patches were 3,862.The results are collected in Table 4 expressed in terms of the main classification metrics (i.e., precision, rejection precision, recall, specificity, accuracy, F1-score). One can note that the combined LIDAR-stereovision system leads to a general improvement of all metrics with an accuracy of 96.5% and F1-score of 98.0%.

Table 4 - Classification results obtained from the single-sensor and combined classifiers for a subset of salient images

|  | LIDAR-based (%) | Stereovision-based (%) | Combined (%) |
| --- | --- | --- | --- |
| Precision | 97.3 | 96.9 | 97.4 |
| Rejection Precision | 83.6 | 82.6 | 90.6 |
| Recall | 97.5 | 97.4 | 98.7 |
| Specificity | 82.7 | 79.7 | 82.8 |
| Accuracy | 95.5 | 95.1 | 96.5 |
| F1-score | 97.4 | 97.2 | 98.0 |

In summary, the combination of stereovision and LIDAR data is useful in that 1) it allows to widen the overall field of view of the perception system with LIDAR and vision working mostly in the short range and long range, respectively, 2) vision can help to overcome limitations of LIDAR, such as sparseness of data and low acquisition frequency, by producing dense maps at relatively high frequency, 3) being less affected by lighting conditions, LIDAR can help to overcome limitations of vision, such as reconstruction errors due to poor lighting conditions, shadows and lack of texture.

6. **Radar-Stereovision combination**

Because of problems of cost and availability of electronic components, radar technology has been long limited to military or aerospace applications. However, it can be successfully used as well for civil applications considering its specific properties: robustness to harsh environmental conditions (dust, rain, fog, snow, day/night cycles, light variations, etc.) and long range measurements.

The K2Pi radar used in this work has been developed by the IRSTEA institute. It is a 24 GHz (K-band) radar, and it is based on frequency modulated continuous wave (FMCW) technology, which is well intended for short and medium distances (up to several hundreds of meters). The radar uses a rotating antenna (horizontal plane) in order to provide each second a panoramic view (360°) of the environment (see Table 2 for more details). During the field experiments, the radar was mounted on the roof of the vehicles (please refer to Fig. 2).

A fan-beam radar, such as the one used in this research, in contrast to pencil-beam counterparts (Reina et al., 2011), offers low vertical resolution. On the other hand, stereovision ensures good resolution to find the boundaries of an object, but range measurements are less accurate than radar. Considering their complementary nature, it is possible to combine the two sensors in order to improve 3D localization of threat obstacles up to about 30 m. A self-supervised scheme is proposed where radar acts as the supervising algorithm that estimates the position of possible hazards spread across the vehicle motion plane. In turn, these 2D range measurements are used to define attention regions in the stereo-generated point cloud that can be analyzed to augment the obstacle information content with 3D geometry and colour data. The radar-stereovision integrated system operates according to a two-step scheme: radar-based 2D multi-target localisation, and radar-stereovision combination for 3D target analysis.

## 6.1. Radar-based 2D multi-target localisation

Radar gives information by radiating a signal in the space and then receiving an echo by the environment. According to the round trip time, this signal is processed and an intensity range profile is generated. As an example, Fig. 7(a) shows the result of the convolution of the sensor beam with the environment in what is usually referred to as a radar image. Note that only the portion of the environment that is seen by both sensors (i.e., common field of view) is considered. However, the radar used in this research is able to provide a 360 deg-panoramic 2D map of the environment with a maximum look-ahead distance of about 100 m. Intensity values above the noise level suggest the presence of an obstacle. Amplitude close to or below the noise level generally corresponds to the absence of objects. The higher the intensity, the larger the cross section or reflectivity of the obstacle encountered, e.g., a large obstacle that is not very reflective will show up with a profile similar to that of a small but highly reflective object. Thus, both types of obstacle will flag the same warning, even though they correspond to different level of danger. This aspect is amplified in fan-beam radar, i.e. with a wide beam angular aperture, where each observation is the result of the integration over a large vertical portion of the environment. As a consequence, the use of fan-beam radar is generally limited to 2D analysis.

Given a 2D radar image, obstacles can be detected through image binarisation and contour and centroid estimation, according to the two following steps:

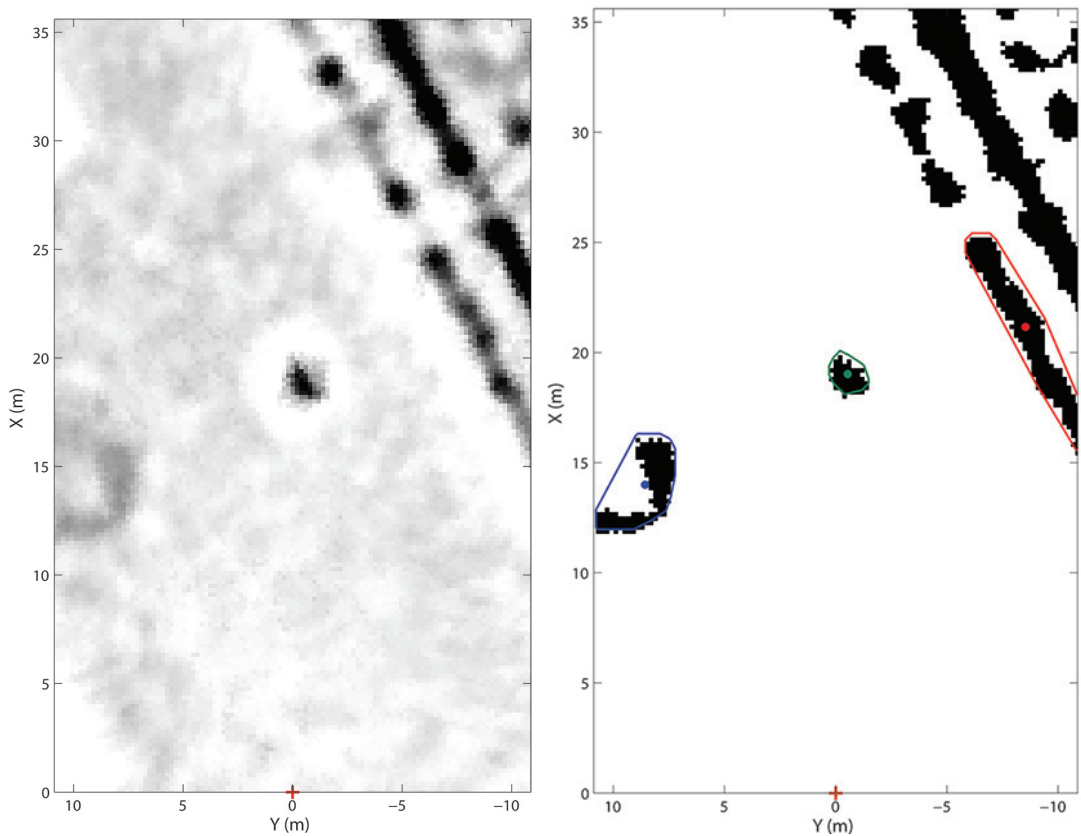

**Fig. 7 -** Radar-based 2D multi-target localisation: a) Raw radar image refereed to the vehicle reference frame. The reference system origin (red cross) indicates the vehicle position. b) Obstacle localization by radar image CFAR binarisation and centroid estimation.

1) *image thresholding*: since in most field applications, noise level in radar image changes both spatially and temporally, it is difficult to set a fixed threshold for target detection. An adaptive approach needs to be pursued, where the threshold level is raised and lowered to maintain a constant probability of false alarm. This is known as constant false alarm rate (CFAR) detection (Brooker, 2005). Fig. 7(b) shows the results of CFAR thresholding. Note that morphological and particle filtering is also applied to enhance image binarisation. Specifically, a morphological opening, i.e., an operation of erosion followed by dilation, is applied to open up touching features and remove isolated background pixels. Then, objects with small areas are filtered out. Finally, a morphological closing, i.e. a succession of dilation and erosion, allows the shape of the obstacles to be constructed by bridging the remaining small gaps.

2) *contour and centroid estimation*: for each radar-detected obstacle, it is possible to define the smallest convex polygon that can contain the region and the relative geometric centroid, as shown in Fig. 7(b). Note that for simplicity only the three closest obstacles to the vehicle are marked.

*6.2. Radar-stereovision combination for target 3D analysis*

Location and contour of a given radar-labelled obstacle is used to define a volume of interest or sub-cloud in the stereo-generated 3D point cloud. First, the sub-cloud is segmented to extract points that actually pertain to the given obstacle. Then, geometric statistics can be obtained for 3D obstacle characterization. In addition, colour data could also be extracted. The stereovision-based obstacle classifier works according to the following steps: sub-cloud extraction and 3D characterisation.

1) *Extraction of the sub-cloud*: 3D points generated by stereovision, whose X and Y coordinates fall into the given obstacle contour are selected. In order to take into account possible misalignments between the two sensors (due for example to calibration errors), points close to the border are also included (i.e., a threshold of 0.5 m was used). As an example, Fig. 8 shows the sub-cloud extracted in the neighbourhood of the closest radar-detected obstacle in Fig. 7(b) (marked by a blue centroid). It corresponds to a chestnut tree at about 14 m to the left of the vehicle.

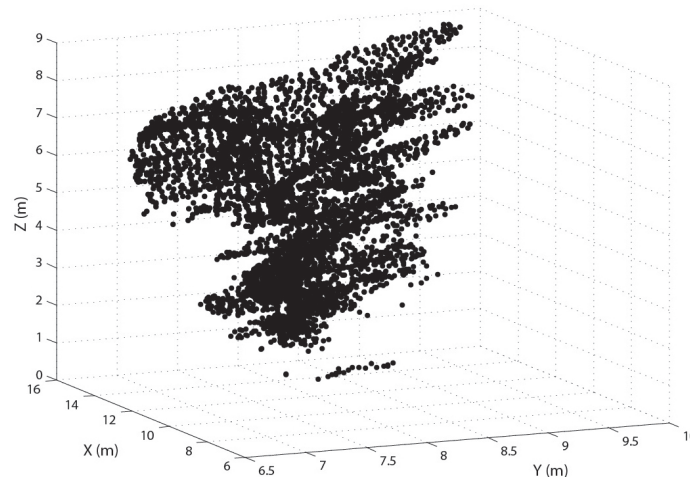

**Fig. 8 -** Radar-based obstacle detection drives extraction of sub-cloud from the 3D point cloud generated by stereovision.

2) *3D obstacle characterisation*: To each radar-labelled obstacle, it will be possible to associate a stereovision-generated 3D point sub-cloud. Following this rationale, stereoscopic 3D geometric and colour properties can be obtained to widen the obstacle information content. As an example, Fig. 9 shows the stereovision sub-clouds corresponding to the radar-labelled obstacles. Each 3D point is plotted using its RGB colour property. The original visual image, with overlaid the projections of the extracted sub-clouds, is also shown, as the background Y-Z plane, whereas the radar map forms the X-Y plane, with green denoting ground and red denoting obstacles. For each obstacle, the maximum height above the ground is reported, as one example of the possible 3D geometric metrics.

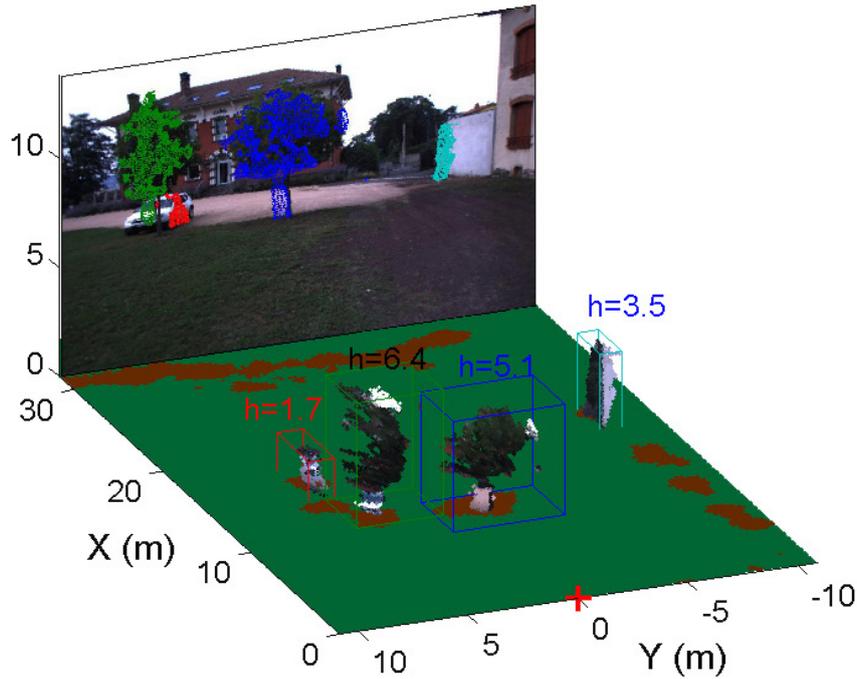

**Fig. 9 -** Radar-based obstacle detection drives extraction of sub-clouds from the 3D point cloud generated by stereovision.

The system was quantitatively evaluated by comparison with "true" obstacle positions that were manually measured in a selected set of frame (i.e. $s_f=62$). Sample results are shown in Fig. 10, where 3D reconstruction provided by stereovision is shown in the neighbourhood of radar-labelled obstacles. In all images, the radar map is reported in the X-Y plane, while the original visual image is displayed in the background Y-Z plane. Stereovision-reconstructed sub-clouds corresponding to radar-labelled obstacles are shown with their respective RGB colour, along with indication of their estimated height. Note that the first 3D reconstruction (upper left image) refers to the radar map of Fig. 7. The accuracy in detecting obstacles was 100% with an error of 0.05 m, obtained as root mean square of the discrepancy along x- and y-axis. The accuracy of the 3D reconstruction by stereovision was quantified by comparing stereo-generated points with a set of points known as the ground truth. The resulting difference defines the 3D measurement error along the three axes, which resulted in an average value of 0.092 m with a standard deviation of 0.062 m.

**7. HDR stereovision-Thermography combination**

In order to overcome common problems with shade and direct sunlight, one possible solution is the use of high dynamic range (HDR) stereovision (Debevec and Malik, 1997). Conventional cameras take images with a limited exposure range, resulting in the loss of detail in bright or dark areas. HDR compensates for this loss of detail by capturing multiple photographs at different exposure levels and combining them to produce an image representative of a broader tonal range. In this research, HDR was performed via software with very fast-synchronised VGA cameras. Specifically, two acA 640-90gc colour cameras (Basler, Ahrensburg, Germany) were used in a stereo setup with 80cm baseline. They are hardware synchronized and take three exposures for high dynamic range images every 40 ms. An example of HDR acquisition is shown in Fig. 11. Pixels are partitioned into 0.6x0.6m cells based on the 3D data. A vector of chromaticity, height from a reference plane and temperature is formed for each cell and used for classification purposes. Fig. 12 shows an overview of the classification scheme. Classification of each cell is based on the distribution of this vector within a cell. The distribution is tested using expectation maximisation (EM) fitting with Gaussian mixture models (GMM). The score is defined using the Bhattacharyya distance of GMMs.

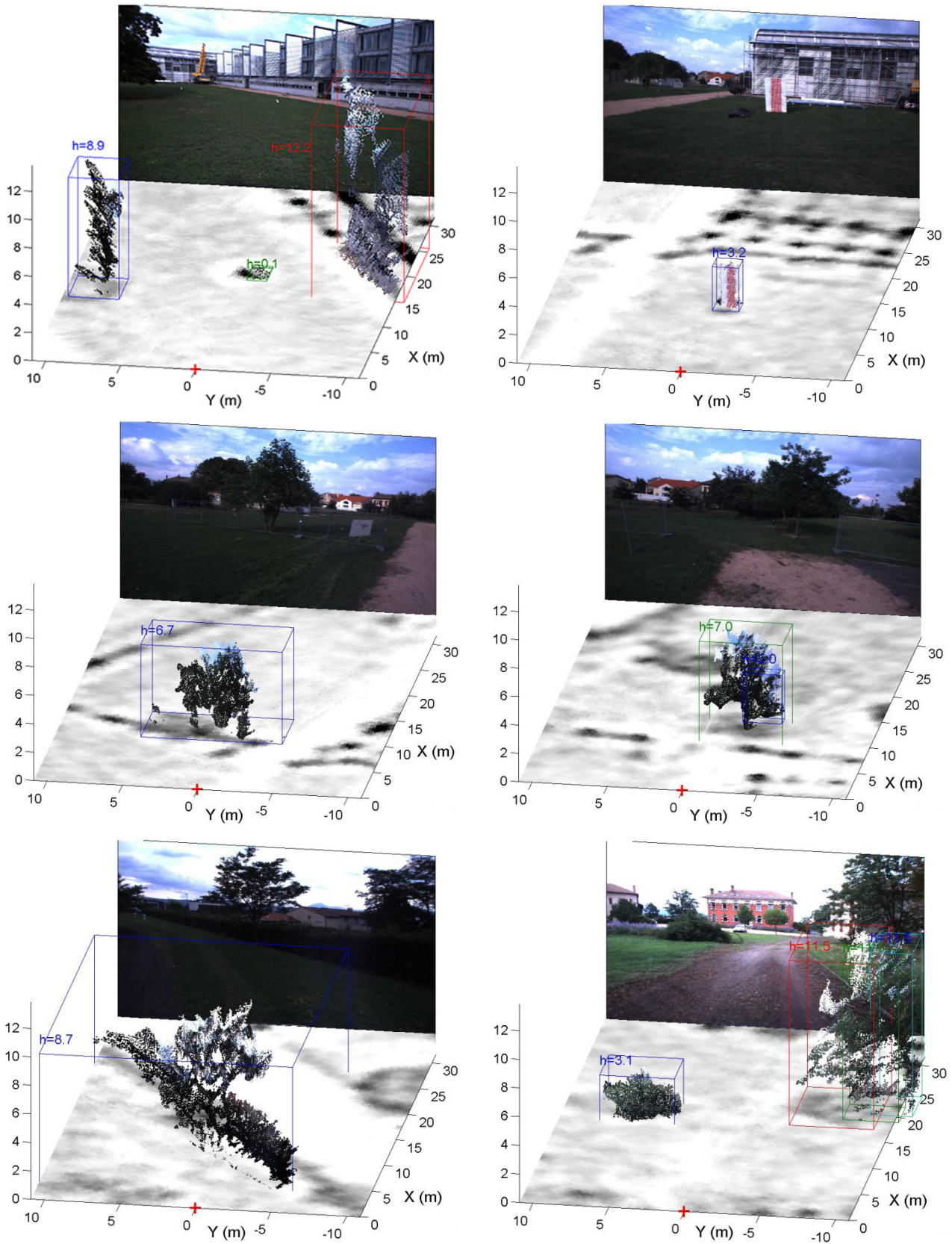

**Fig. 10 -** 3D reconstruction of radar-labelled obstacles using stereovision.

An A615 infrared camera (Flir, Andover, MA, USA) is also co-located with the stereovision rig for HDR-thermography combination. 3D reconstruction is done real-time using CUDA acceleration and the thermal image is warped to corresponding colour pixels using two-pass dynamic programming (Kim et al., 2005).

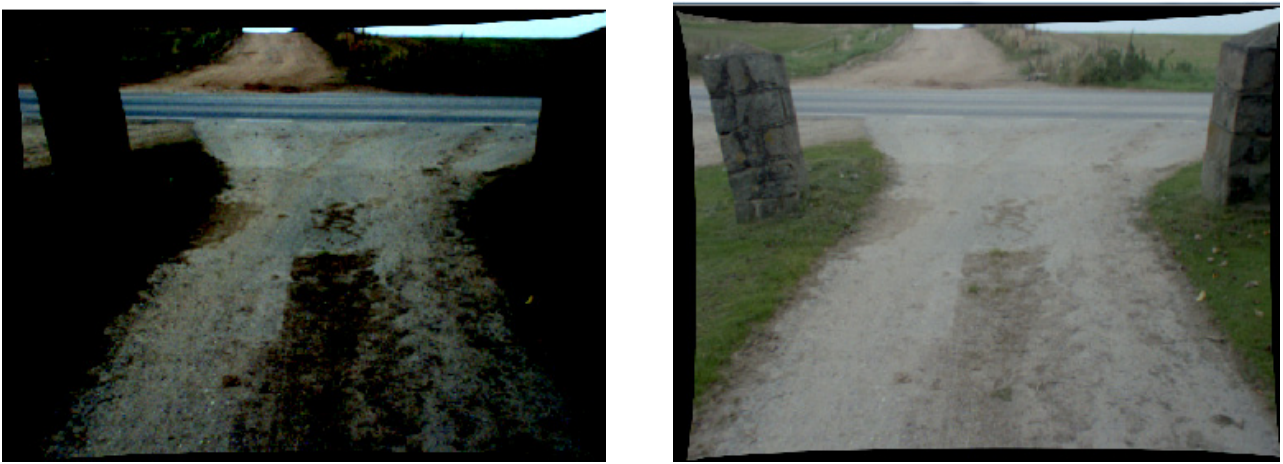

**Fig. 11 -** Left: one exposure. Right: HDR output

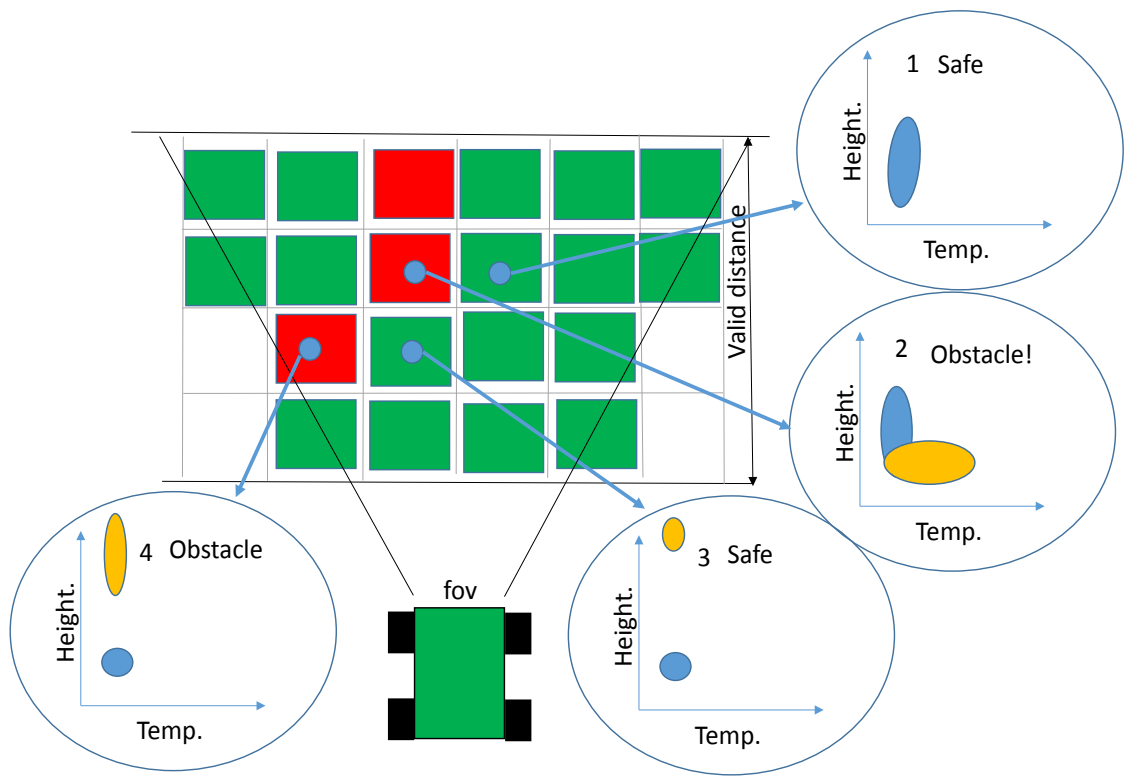

**Fig. 12 -** The FOV in front of the tractor was divided into a grid and the classification was based on the "distribution of feature values", rather than absolute values. Cells behind an obstacle were marked as not traversable, too. Four examples of cell distribution are shown in 2D plots for Height and Temperature: 1. Typical tall crop situation. Many heights exist. It is safe. 2. In the same tall crops, there is now a small animal: STOP! 3. Flat ground and high hanging branch above the tractor. It is safe. 4. Now the branches are too low: STOP!

Sample results are collected in Fig. 13. The first row shows the tractor approaching a main road. The stones on each side of the entrance are classified as obstacles while the slight inclination of the road is seen as traversable. Suddenly, a car passes (second row), and it is flagged by the system. The position of the car matches in the stereovision images and thermal image. Third to sixth row show the most interesting cases, where a tall crop can be classified as traversable. In the third row, the left cell is selected (cyan = traversable + selected) and it is traversable. To the right the GMM space is shown, as tall vertical ellipsoids. The middle cell is red (obstacle) and 4th row selects that cell. Now, the GMM shows two unique components. The RGB image shows maize, whereas the thermal image indicates a person behind the maize. 5th and 6th row shows tall grass that is traversable, but then a person appears and it becomes an obstacle.

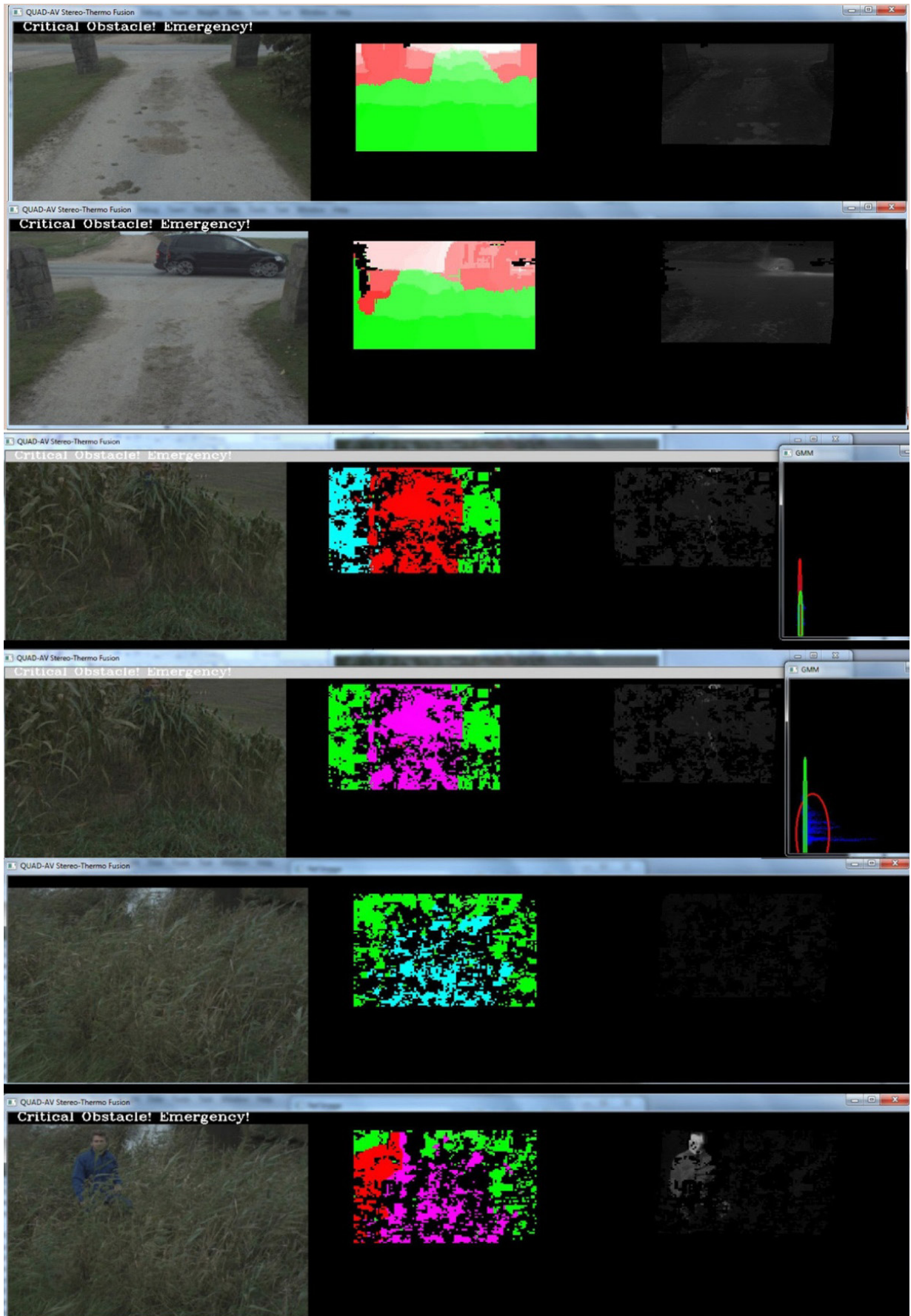

**Fig. 13 -** Results shown in the following way: RGB Image / Classified Cells / Thermal Image that has a smaller FOV that limited the amount of annotated cells. Green cells: traversable. Red: not traversable. Purple/Cyan: selected cell for showing the GMM classes (when applicable).

The data set used for testing the algorithm for HDR stereovision-thermography combination consisted of 32 different situations that were hand-labelled by visual inspection. The scenes included open fields, lakes, puddles, humans, maize, tall grass, hedges, game birds, buildings, vehicles, fences, ramps, road crossings, hills, large stones, and gravel roads. A quantitative evaluation of the classifier was obtained by measuring the main statistical measures of performance, as collected in Table 5. All metrics were high attesting to the feasibility of the proposed approach. The main reasons for false positives and false negatives are briefly discussed in Table 6 along with examples of false negatives shown in Fig. 14.

Analysis of the algorithm's response showed that the significant features were the distributions of height from ground plane and temperature. Obstacles were detected well with chromaticity weight zeroed and equal weight on height and temperature. The temperature was an important feature for detecting living things and vehicles, but also stones and poles due to their different emissivity. It made it easy to detect people hiding 5 m inside tall maize, and crouching in tall grass. The height made it possible to see fences, half meter tall concrete barriers versus valid ramps. The depth resolution was limited at a distance, so small fences and hedges were not detected until within 15 m.

**Table 5 -** Statistical measures of performance, as obtained by the HDR stereovision-thermography combination

|  | HDR-thermography combination (%) |
|---|---|
| Precision | 99.3 |
| Rejection Precision | 95.3 |
| Recall | 97.4 |
| Specificity | 98.8 |
| Accuracy | 97.9 |
| F1-score | 98.4 |

**Table 6 -** Main reasons for false positives and false negatives in the HDR stereovision-thermography combination

| False positives | |
|---|---|
| **Reasons** | **Discussion** |
| Edge of image | Cells can be invalidated at the edge of the image to avoid this issue. |
| Cell is only partially negative | The larger cell size seems to improve the rejection precision on large surfaces, so to avoid issues around class boundaries, the boundary cells should be double checked and partitioned. |
| **False negatives** | |
| **Reasons** | **Discussion** |
| Water puddle | This is a feature that should stop the vehicle in the presence of a lake, but it cannot see if it is a lake or a puddle. Fusion with e.g. radar, GIS data, geofencing could be a solution. Refer to Fig. 14(a) for an example. |
| Illusion of negative cliff | Then approaching the top of a steep hill or higher ground, when looking over the top you cannot see how steep it is from that vantage point. But it will be apparent when you get closer. Also radar and GIS data could be used to improve this. Refer to Fig. 14 (b) for an example. |
| A jacket still warm left in the grass | This is the main issue with thermal obstacle detection. Should be a priority in future work. Refer to Fig. 14(c) for an example. |
| Tall grass mistaken for maize | Under the condition where tall grass is trained as traversable but maize is not, the current implementation lacks the spectral- and texture signatures of the different types of vegetation to accurately tell the difference. |
| Cell is only partially positive | Cells at the boundary of classes should be partitioned to fix this issue. |

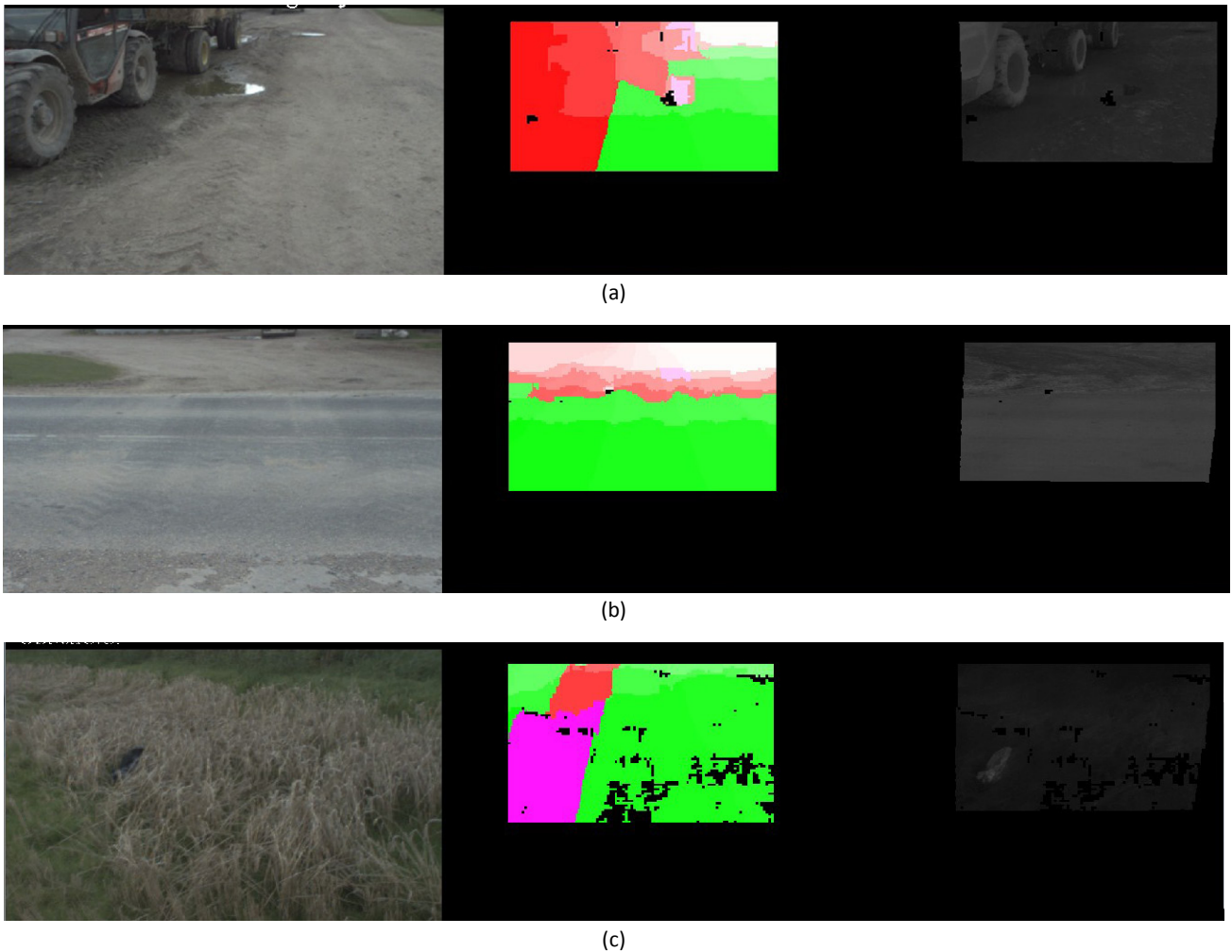

**Fig. 14** - Examples of false negatives: a water puddle (a), Illusion of negative cliff (b), a warm jacket in the grass (c).

Fusion of stereovision and thermography potentially provides a powerful obstacle detector in agricultural settings, where the traversable ground cannot be assumed to be flat. Training can be done by driving the vehicle through traversable paths and crops. However, a vehicle can be pre-initialised with certain knowledge about certain obstacles. Certain classes can also be assigned higher priorities, such as humans.

## 8. Conclusions and future work

The development of a multi-sensor platform for agricultural vehicles in crop fields was presented. Four sensor technologies and their combinations were evaluated: stereovision, LIDAR, radar, and thermography. Specifically, the problem of reliably detecting obstacles and traversable ground surface was addressed, and innovative methods were developed and validated in the field.

One major concern for the development of such methods was the lack of *a priori* knowledge about the environment and the high variability of the situations that may be encountered, including varying terrain properties and changing illumination conditions. To address these issues, an online self-learning framework for traversable ground detection based on geometric features was firstly presented, where the terrain model is automatically initialised at the beginning of the vehicle's operation and progressively updated online. Self-supervised systems reduce or eliminate the need for hand-labelled training data, thus gaining flexibility in unknown environments. The proposed approach was demonstrated for stereovision data acquired by a multi-baseline system, which enables the vehicle to perform accurate 3D scene reconstruction from near range up to several meters away.

Being passive devices, cameras are affected by environmental factors, such as lighting conditions. In addition, stereovision-generated maps may be corrupted by poor reconstruction due for instance to textureless areas or occlusions. Hence, visual input must be combined with information from other complementary sensors. The integration of stereovision with LIDAR, radar and thermal imaging respectively was investigated. Specifically, a high-level fusion approach to combine a LIDAR-based classifier and a stereovision-based classifier to detect traversable ground was proposed. It was shown that vision can help to overcome limitations of LIDAR, such as sparseness of data and low acquisition frequency, by producing dense maps at relatively high frequency, whereas LIDAR can help to overcome limitations of vision, such as reconstruction errors due to poor lighting conditions, shadows and lack of texture.

An integrated radar-stereovision system was also presented for 3D obstacle detection. It features a self-supervised scheme where radar serves as the supervising algorithm detecting 2D coordinates of possible hazards spread across the vehicle motion plane. Radar measurements are then used to define attention regions in the stereovision-generated point cloud, so as the obstacle information content can be augmented with 3D geometry and colour data.

Finally, a combination of thermal imaging and HDR stereovision was proposed. It was shown that 3D stereovision measurements (i.e., height from a reference plane) and temperature provide significant features for detection of living beings and other obstacles like vehicles, stones and poles.

The investment cost of a complete system may be high, but our modularity means that the farmer only has to pay for those sensors which are useful to his particular crop, terrain, and climate. The aim for our system is to produce considerable savings due to early detection and treatment of problems. For example, if the reduction in machine damage due to collisions is taken into account, and especially if farmers can get a reduction in insurance premiums due to the expected reduction in rollover and damage to third parties, then the savings also become significant.

Further research will consider methods:

- to address issues related to "transitory warm" objects, which are erroneously flagged by thermography.
- to create persistent perception based on sensor data fusion for sensing in adverse visibility conditions;
- to increase the safety by robust perception, which is also able to work after a failure has occurred (graceful degradation);
- to improve ambient awareness (traversable terrain, obstacles, crop assessment, etc.) from the fused sensor data using methods of semantic classification;
- to enable a team of agricultural vehicles for cooperative interaction (master-slave setup) as well as for interaction with the human operator (HRI).

**ACKNOWLEDGMENT**

The financial support of the FP7 ERA-NET ICT-AGRI through the grants Ambient Awareness for Autonomous Agricultural Vehicles (QUAD-AV) and Simultaneous Safety and Surveying for Collaborative Agricultural Vehicles (S3-CAV) is gratefully acknowledged.